\def\year{2021}\relax
\documentclass[letterpaper]{article} 
\usepackage[switch]{lineno}
\usepackage{aaai21}  
\usepackage{times}  
\usepackage{helvet} 
\usepackage{courier}  
\usepackage[hyphens]{url}  
\usepackage{graphicx} 
\def\UrlFont{\rm}  
\usepackage{natbib}  
\usepackage{caption} 
\usepackage[ruled, linesnumbered]{algorithm2e}
\usepackage[table,xcdraw]{xcolor}
\usepackage{amsmath}

\newcommand{\toolname}{LGGA\xspace}
\newcommand{\dataefficency}{61.9\%\xspace}
\newcommand{\uniquesolved}{30.0\%\xspace}
\title{ Logic Guided Genetic Algorithms}
\author{
    Dhananjay Ashok\textsuperscript{\rm 1},
    Joseph Scott\textsuperscript{\rm 2},
    Sebastian J. Wetzel\textsuperscript{\rm 3},
    Maysum Panju\textsuperscript{\rm 2},
    and Vijay Ganesh\textsuperscript{\rm 2}
    \\
}
\affiliations{

    \textsuperscript{\rm 1}University of Toronto, Canada\\
    
    \textsuperscript{\rm 2}University of Waterloo, Canada\\
    \textsuperscript{\rm 3}Perimeter Institute for Theoretical Physics, Canada\\
    dhananjay.ashok@mail.utoronto.ca, \{joseph.scott, mhpanju, vijay.ganesh\}@uwaterloo.ca, swetzel@perimeterinstitute.ca
}

\begin{document}

\maketitle

\begin{abstract}
We present a novel Auxiliary Truth enhanced Genetic Algorithm (GA) that uses logical or mathematical constraints as a means of data augmentation as well as to compute loss (in conjunction with the traditional MSE), with the aim of increasing both data efficiency and accuracy of symbolic regression (SR) algorithms. Our method, logic-guided genetic algorithm (LGGA), takes as input a set of labelled data points and {\it auxiliary truths} (AT) (mathematical facts known a priori about the unknown function the regressor aims to learn) and outputs a specially generated and curated dataset that can be used with any SR method. Three key insights underpin our method: first, SR users often know simple ATs about the function they are trying to learn. Second, whenever an SR system produces a candidate equation inconsistent with these ATs, we can compute a counterexample to prove the inconsistency, and further, this counterexample may be used to augment the dataset and fed back to the SR system in a corrective feedback loop. Third, the value addition of these ATs is that their use in both the loss function and the data augmentation process leads to better rates of convergence, accuracy, and data efficiency. We evaluate \toolname against state-of-the-art SR tools, namely, Eureqa and TuringBot on 16 physics equations from ``The Feynman Lectures on Physics" book. We find that using these SR tools in conjunction with \toolname results in them solving up to \uniquesolved more equations, needing only a fraction of the amount of data compared to the same tool without \toolname, i.e., resulting in up to a \dataefficency improvement in data efficiency.
\end{abstract}

\section{Introduction}
\label{sec:intro}
Symbolic Regression (SR) is a well studied problem in artificial intelligence (AI), where the objective is to find a mathematical or symbolic expression, over a fixed set of mathematical symbols (the alphabet), that approximates an as-yet-unknown function $f(x_1,x_2...x_n) = y$ given only a training dataset~\cite{koza1992genetic}. In principle, an SR tool can automate the process of discovering equations or formulae from experimental data, a task which takes scientists and engineers a significant amount of time and effort. The importance of the field can be seen in its prominent use in modeling complex dynamic systems in an interpretable way~\cite{billard2002symbolic}, as well as the many industrial-strength SR tools, such as Eureqa~\cite{schmidt2009distilling} and TuringBot~\cite{turingbot}, available to engineering companies. Both these tools are highly engineered, robust, with broad applicability in many domains, and consequently pose a challenge to improve upon further.

There is a general consensus that SR is an NP-hard optimization problem~\cite{udrescu2020ai,lu2016using,towfighi2020symbolic},  leading to the prominent use of genetic algorithms (GAs) which are well suited to searching intractably large problem spaces~\cite{nicheparetogp}. We refer the reader to the seminal books~\cite{koza1992genetic,pal1996genetic} and papers~\cite{gptips} that showcase the use of GAs in the SR setting. 

Recently, several Deep Neural Network (DNN) based methods have been proposed to address the SR problem, the most successful among them being AI Feynman 1.0 and 2.0~\cite{udrescu2020ai,udrescu2020ai2}. Briefly, both versions of AI Feynman use a combination of specialized properties common in physics equations (e.g., additive and multiplicative separability) and use DNNs to break down the problem into several smaller SR problems which can then be tractably solved.

\begin{table*}[t]
\begin{tabular}{|llll|}
\rowcolor[HTML]{AEAAAA} 
\hline
S.No & Name           & Equation                                    & Auxiliary Truths                                                                                               \\
1    & Resistance     & $\frac{r_1r_2}{r_1+r_2}$               & $(SZ)_{r_1, r_2}$, output is smaller than inputs                                                       \\
2    & Snell          & $\frac{\sin(i)}{\sin(r)}$               & $f(x_1, x_2) = \frac{1}{f(x_2, x_1)}$ if neither the LHS or the RHS is 0.                            \\
3    & Coulomb        & $k\frac{q_1q_2}{r^2}$                       & $(SZ)_{q_1, q_2}$, output positive iff $q_1, q_2$ same sign                                              \\
4    & Reflection     & $|\frac{n_1-n_2}{n_1+n_2}|^2$               & Range is $(0,1), S_{n_1, n_2}$                                                                        \\
5    & Gas            & $\frac{PV}{nT}$                             & $(SZ)_{P, V}, S_{n, T}$                                                                                  \\
6    & Distance       & $\sqrt{(x_1-x_0)^2+(y_1-y_0)^2}$            & $S_{x_0, x_1}, S_{y_0, y_1}$; if all of $x_0,x_1,   y_0$  are  0 then output $y_1$ \\
7    & Normal         & $\frac{e^{-x^2}}{2\pi}$                     & $S_{x, -x}$; $x = 0 \Leftrightarrow 0.1591549$                                                        \\
8    & Dot            & $x_1y_1+x_2y_2+x_3y_3$                      & If all $x$ or all $y$ are 0 then output 0; if all inputs same, output is $3x_{1}^2$                             \\
9    & Field          & $q(Ef+Bv\sin(\theta))$                      & $Z_{q}, S_{B,v}$                                                                                       \\
10   & Potential      & $\frac{Gm_1m_2}{\frac{1}{r2}-\frac{1}{r1}}$ & $(SZ)_{m_1, m_2}$                                                                                       \\
11   & Centre of Mass & $\frac{m_1r_1+m_2r_2}{m_1+m_2}$             & If all $r$ are 0 then output 0; $m_1=0\Rightarrow$output $r_2$                                     \\
12   & Momentum       & $mrv\sin\theta$                             & $(SZ)_{m, r, v}$                                                                                        \\
13   & Mass           & $\frac{m_0}{1-\frac{v}{c}}$                 & $Z_{m_0}$                                                                                             \\
14   & Heat           & $\frac{1}{\gamma-1}prV$                     & $(SZ)_{pr, v}$                                                                                          \\
15   & Boyle          & $nk_bT\ln(V_2/V_1)$                         & $(SZ)_{n, k\_b, T}$                                                                                     \\
16   & Flow           & $\sqrt{\frac{pr\gamma}{\rho}}$              & $(SZ)_{pr, \gamma}$, if all values 1 output 1                                                          \\
\hline
\end{tabular}
\caption{Complete list of equations from the Feynman Lectures in Physics book used in our experiments, along with associated auxiliary truths. The expression $S_{x_1, x_2, ..x_n}$ denotes that the unknown function $f$ is  {\bf symmetric} in all pairwise arguments $x_1,x_2,...,x_n$. The expression $Z_{x_1, x_2, ...}$ denotes the {\bf zero condition}, that is, if any of the values arguments is 0 then the output of $f$ is 0 as well. The expression $(SZ)_{x_1, x_2, ..}$ denotes the combination of both $S_{x_1, x_2, ..}$ and $Z_{x_1, x_2, ...}$, that is, the function $f$ respects both symmetry and the zero condition on $x_1,...,x_n$.}
\label{table:equation_list}
\end{table*}

While the methods mentioned above represent an impressive advance in addressing the SR problem over the last several decades, to the best of our knowledge none of these methods leverage domain-specific knowledge or auxiliary truths (ATs) --- mathematical facts known a priori about the unknown function that the regressor aims to learn. These ATs are typically simple domain-specific properties of the unknown function $f$ that SR users are likely to know and can significantly enable SR methods to be more data efficient. Further, in recent years, we have seen an increasing effort by researchers to bring together the two pillars of AI, namely, machine learning (ML) and mathematical logic, with the aim of solving problems that neither approach by itself can feasibly solve~\cite{belle2020symbolic}.

These two observations serve as the impetus for our work on a constraint enhanced GA that uses logical or mathematical constraints as a vehicle for data augmentation as well as to compute loss in conjunction with the traditional mean square error (MSE), with the goal of increasing the data efficiency and accuracy of SR tools. Our method, logic guided genetic algorithms (LGGA), takes as input a set of labelled data points, ATs, and an alphabet, and outputs both an equation prediction and a specially generated and curated dataset that can be used with any SR method. 

\vspace{0.1cm}
{\bf The following three key insights underpin our approach:} First, the users of SR systems often have some domain-specific knowledge or ATs about the unknown function that they could use to make the SR system more data efficient and effective than otherwise. Often these ATs are elementary properties of said unknown function. Second, the fundamental property of ATs is that any purported function that fits the input data must be consistent with them. That is, whenever an SR system outputs a symbolic expression, and it happens to be inconsistent with the ATs (i.e., a counterexample can be computed), then such counterexamples may be used to augment the dataset and fed back to the SR system in a corrective feedback loop. In addition to such data augmentation, ATs can be used as part of a (weighted) loss function of a GA. Finally, if used appropriately, this kind of domain-specific knowledge can be beneficial in guiding SR systems to converge efficiently to a more accurate symbolic representation of unknown functions than otherwise.

\noindent{\bf Problem Statement:} In brief, the problem we aim to solve is the following: How can the user of an SR system leverage their domain-specific mathematical knowledge (i.e., ATs), with the aim of making SR more data efficient and effective? (By data efficient we mean that an SR system with ATs may need fewer data points than the same system without ATs. By the term {\it effective} we mean that an SR system with ATs may be able to learn a function while the one without ATs would fail to (within a reasonable timeout).)

\subsection{Contributions.}

\noindent{\bf 1. The LGGA Algorithm:} We present a novel AT-enhanced GA aimed at the symbolic regression problem, dubbed logic guided genetic algorithm (\toolname), where ATs are typically elementary properties of the unknown function $f$. The ATs are used in two ways in our algorithm: first, as part of a (weighted) loss function along with MSE, and second, as a way of generating new data that can enable an SR system to learn a more accurate approximation of the unknown function in significantly more data-efficient manner (the data augmentation feature of \toolname). (See Section~Logic Guided Genetic Algorithms)

\noindent{\bf 2. Extensive Experimental Evaluation:} To test our LGGA method, we augment three different state-of-the-art SR methods, namely, Eureqa, TuringBot, and AI Feynman 2.0, with the {\it LGGA process}. The goal is to perform an apple-to-apple comparison of these industrial-strength SR methods with and without LGGA. As a benchmark, we selected 16 physics equations from the ``Feynman Lectures on Physics" book~\cite{feynman2011feynman}. Physics equations are an excellent way to benchmark and debug SR methods since they are well known and can be easily used to generate data. During the evaluation, the SR methods are only given data and ATs as inputs, and are then required to output a symbolic expression that is semantically equivalent to (or an approximation of) the equation that corresponds to the data (and is as-yet-unknown to the SR method). We show that the LGGA-enhanced SR methods are significantly more data efficient than their non-LGGA counterparts. In some cases, the LGGA-enhanced SR methods can learn a function, while the non-LGGA counterpart cannot learn even with very large timeouts. Further, the LGGA versions produce a more accurate approximation of the unknown function than the non-LGGA counterparts. More precisely, we find that using these SR tools in conjunction with \toolname results in them solving up to \uniquesolved more equations. Finally, \toolname-enhanced Eureqa and TuringBot only need a fraction of the amount of data needed relative to their non-LGGA counterparts, i.e., they demonstrate up to a \dataefficency improvement in data efficiency. Of the three pairs of methods we compared, The LGGA-enhanced version of TuringBot seems to be the most efficient, accurate, and robust. (See Section~Results)

\section{Motivation for Auxiliary Truths (ATs) in SR}
\label{sec:auxiliarytruth}
In this section, we motivate how and why ATs can be an effective way of both enhancing the power and augmenting data for any GA (and more generally, any SR) algorithm. Traditional SR systems take as input a dataset of the form ${\bf X}, {\bf y}$ and a set of symbols $\Sigma$ (alphabet), and output a candidate symbolic expression $\widehat{f}(x_1,\ldots,x_n)=y$ over the input alphabet such that $\widehat{f}$ fits the data. In \toolname, we have a third input, the ATs. The full list of equations and ATs we work with in this paper is given in Table~\ref{table:equation_list}.

\noindent{\textbf{Auxiliary Truths:}} We define the term ATs as mathematical expressions that capture domain-specific knowledge or simple properties of an unknown function $f$ to be learnt. It goes without saying that auxiliary truths are not the same as the unknown function $f$ in question, but rather are relatively simple properties of $f$ that can be useful in making SR systems more efficient than otherwise.

\vspace{0.05cm}
\noindent{\textbf{Example of ATs:}} Consider the following example of a function $R = \frac{r_1r_2}{r_1+r_2}$ (Parallel Resistance) that a user of an SR tool may want to learn from data. A physicist could rely on their domain knowledge about resistors and infer that if one of the resistors in a pair of parallel resistors has a resistance of zero, the combined resistance is zero. Further, exchanging the resistors would yield the same combined resistance. The user would quickly infer the following two simple ATs.

\begin{itemize}
    \item Symmetry: $\forall r_1,r_2. |R(r_1,r_2) - R(r_1, r_2)| = 0$
    \item Zero Conditions: $(r_1=0\vee r_2=0)\Rightarrow R(r_1,r_2)=0$
\end{itemize}

Observe that the SR user doesn't have to know the actual equation $R = \frac{r_1r_2}{r_1+r_2}$ in order to infer the above properties.

\vspace{0.05cm}
\noindent{\textbf{Properties of ATs:}} Importantly, all of the ATs or function properties discussed and used in this paper satisfy the following criteria:
\begin{enumerate}
    \item A typical SR user may very well know ATs without knowing the actual equation/function she wants to learn from data. For example, laboratory experiments aimed at collecting data could reveal symmetry or the value of a function at 0, even if the user has no idea of what the final symbolic form the equation may take.
    
    \item One can evaluate whether an arbitrary unknown function $f$ is consistent with ATs without using labelled data or knowing the target function $f$
    
    \item A crucial property of any AT we consider is that it must be {\it  consistent} with any candidate function that an SR system outputs. That is, any candidate learnt function or symbolic expression is trivially incorrect if it is not consistent with the input ATs (we do assume that SR users happen to know the correct ATs associated with the unknown function). Further, whenever an SR system produces a candidate symbolic expression that fits the input data and if such an expression happens to be inconsistent with the ATs (i.e., a counterexample or an ``adversarial datapoint" can be computed), not only do we \textbf{receive a signal that the candidate function is incorrect} which can be used as part of a loss function, but we also can \textbf{use the counterexample to augment the dataset} and feed it back to the SR system in a corrective feedback loop~\footnote{In general, consistency checking of formulas from a suitable fragment of mathematics reduces to the satisfiability problem for the said fragment. Given that the complexity of the satisfiability problem for various fragments of math can be very high, e.g., NP-complete or even undecidable, we limit our \toolname tool to ``formula evaluation over boundary conditions or given data". Fortunately, such formula evaluation is usually computationally very cheap, while simultaneously can have a profound effect on the efficacy of an SR system.}.)
\end{enumerate}

\begin{figure}[t]
\centering
\includegraphics[width=\linewidth]{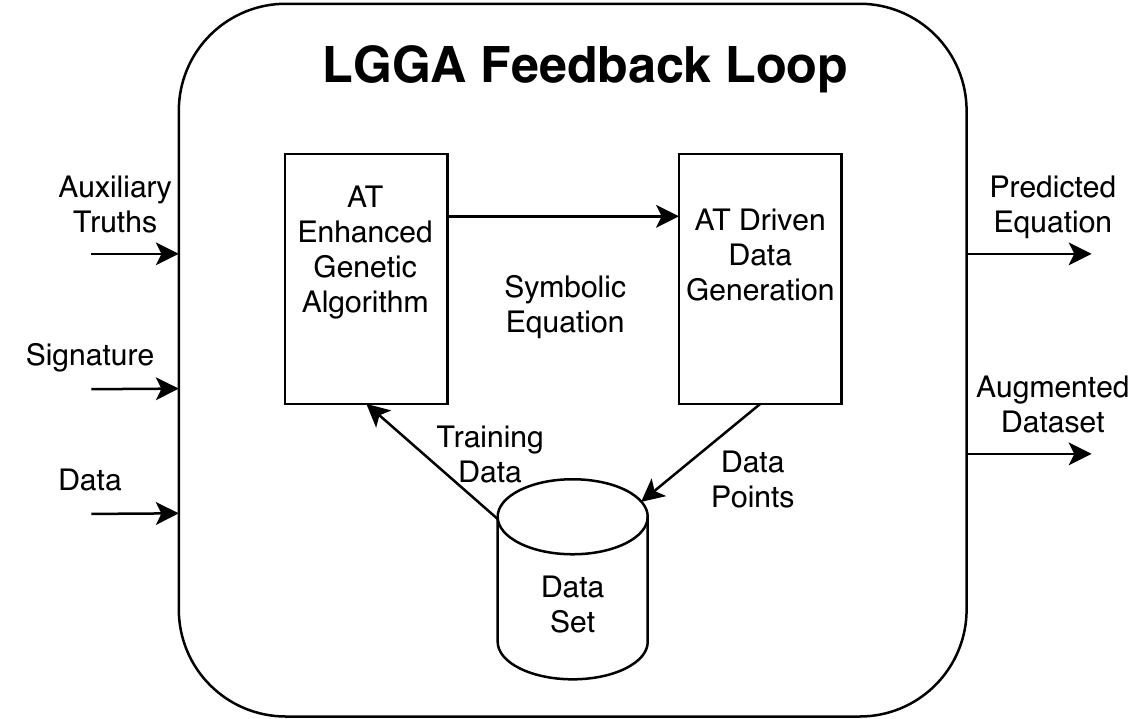}
\caption{Architecture Diagram for \toolname. The data, ATs and alphabet (allowed operations) are given as inputs to the AT-constraint enhanced genetic algorithm. This outputs an equation to the AT-enhanced data generation unit, where the ATs are used to generate new data points which are then added back into the dataset. This feedback loop continues until termination, giving an augmented dataset.}
\label{fig:arch}
\end{figure}

\noindent{\textbf{The Motivation for ATs as part of the \toolname Loss Function:}} Once again, consider the $R = \frac{r_1r_2}{r_1+r_2}$ equation with the ATs stated above. When we attempted to discover this equation from data using a simple genetic algorithm, our prediction after four generations was $R = r_1 + r_2$. This prediction, however, breaks our "zero condition" AT, and so when we rerun the very same algorithm, but use this AT along with MSE as part of a weighted loss function, we see that after four generations, the algorithm converges to $R = \frac{r_1r_2}{r_1+r_2}$. This example motivates the idea (that will be explained in detail in the subsequent section) that a loss function that penalizes a candidate equation for violating known ATs and MSE is more powerful than a loss function that focuses on MSE alone.

\noindent{\textbf{The Motivation for ATs in Data Augmentation:}} One of the most common drawbacks of powerful SR systems is their propensity to overfit the input data, often ending up with a complex polynomial function approximation of high degree. This problem is most pronounced when there is limited data, and the model quickly reduces to overfitting instead of finding the true equation. To illustrate this point, we use the industrial-strength Eureqa tool. When supplied with 15 random data points for the gas equation $n = PV/RT$, the Eureqa tool successfully finds the target equation. However, once the number of random data points is reduced to 6, Eureqa begins to fail and produce the following complex overfitting approximation instead: 
\begin{equation}
206+3.41V+1230\cos(V)+138\cos(V)^2-V^2\cos(V)...-T^2     
\label{eq:eureqa_wrong}
\end{equation}

Notice how there is a constant term in the above equation, and hence this violates the constraint that whenever $P$ or $V$ are 0, then the result is 0. The case for data augmentation using ATs is motivated by the fact that often complicated approximations may be good at fitting limited data points, but they are deficient in being consistent with ATs. This example suggests that even an industrial-strength tool like Eureqa can suffer when the quality of the data it is given is relatively low, and there is a strong case to be made for the use of \toolname's data generation method as a pre-processing step to give the final model a smaller but richer set of data points which would add value in cases where there is a paucity of useful ``adversarial" data.

\section{Logic Guided Genetic Algorithms}
\label{sec:lgga}
\begin{table*}[ht]
\centering
\begin{tabular}{|l|
>{\columncolor[HTML]{FFFFFF}}l 
>{\columncolor[HTML]{FFFFFF}}l l|
>{\columncolor[HTML]{FFFFFF}}l 
>{\columncolor[HTML]{FFFFFF}}l l|}
\rowcolor[HTML]{AEAAAA}
\hline
\multicolumn{1}{|c}{Equations}         & \multicolumn{3}{c}{Minimum Data Points Needed for Eureqa}                                                          & \multicolumn{3}{c|}{Minimum Data Points Needed for TuringBot}                                                       \\
\hline
\rowcolor[HTML]{AEAAAA}
Target Equation                         & LGGA & No LGGA & DE \% & LGGA & No LGGA & DE \% \\
$\frac{r_1r_2}{r_1+r_2}$                                                  & $8\pm 2$                            & $21\pm 3$                              & 62                                         & $1\pm 0$                            & $1\pm 0$                               & 0                                          \\
$\frac{\sin(i)}{\sin(r)}$                                                 & $6\pm 1$                            & $14 \pm 2$                              & 58                                      & $6 \pm 1$                            & $10 \pm 2$                              & 40                                         \\
$\frac{Kq_1q_2}{r^{2}}$                              & $6\pm 1$                            & $14\pm 2$                              & 58                                      & $2\pm 0$                            & $2\pm 0$                               & 0                                          \\
\textbf{$|\frac{{n_1-n_2}}{{n_1+n_2}}|^{2}$}                          & \textbf{$200\pm 40$}                          & \textbf{NoDisc}                               & \textbf{Disc}                                       & \textbf{$600\pm 100$}                          & \textbf{NoDisc}                               & \textbf{Disc}                                       \\
$\frac{PV}{nT}$                                                         & $6\pm 3$                            & $15\pm 4$                              & $60$                                         & $4\pm 0$                            & $4\pm 0$                               & 0                                          \\
\textbf{$\sqrt{(x_1-x_0)^{2} + (y_1-y_0)^{2}}$} & \textbf{$200\pm 100$}                          & \textbf{NoDisc}                               & \textbf{Disc}                                       & $2000\pm 800$                            & \textbf{NoDisc}                               & \textbf{Disc}                                          \\
\textbf{$\frac{e^{-x^{2}}}{2\pi}$}                             & \textbf{$190\pm 100$}                          & \textbf{NoDisc}                               & \textbf{Disc}                                       & $300\pm 50$                          & $800\pm 200$                             & 62.5                                       \\
$x_1y_1 + x_2y_2 +x_3y_3$                                             & U                            & U                               & U                                          & $10\pm 3$                           & $20\pm 2$                              & 50                                         \\
$qEf + qBv\sin(\theta)$                                         & U                            & U                               & U                                          & $65\pm 10$                           & $120 \pm 25$                             & 45                                        \\
$\frac{Gm_1m_2}{\frac{1}{r_2}-\frac{1}{r_1}}$                                           & $13\pm 5$                           & $23\pm 3$                              & 45                                         & $10\pm 5$                           & $25\pm 7$                              & 60                                         \\
$\frac{m_1r_1+m_2r_2}{m_1+m_2}$                                         & $22\pm 3$                           & $38\pm 5$                              & 43                                       & $7\pm 0$                            & $14\pm 3$                              & 50                                      \\
$mrv\sin(\theta)$                                              & $23 \pm 6$                           & $48\pm 7$                              &  52                                     & $11\pm3$                           & $20\pm 5$                              & 45                                         \\
$\frac{m_0}{1-\frac{v}{c}}$                                                  & $34\pm 2$                           & $41\pm 3$                              &  20                                      & $5\pm 0$                            & $5\pm 0$                               & 0                                          \\
$\frac{Vpr}{\gamma-1}$                                              & $40\pm 10$                           & $65\pm 5$                              & 40                                      & $5\pm 0$                            & $6\pm 0$                               & 16.66                                      \\
$nkbT\ln(\frac{V_2}{V_1})$                                              & U                            & U                               & U                              & $300\pm 50$                          & $720\pm 100$                             & 60                                      \\
$\sqrt{\frac{pr\gamma}{\rho}}$                                            & $23\pm 5$                           & $55\pm 6$                              & 59                                      & $1300\pm 750$                           & $2000\pm 700 $                              & 35                                         \\
$\frac{gqB}{2m}$                                                 & $15\pm 4$                           & $40\pm 3$                              & 62.5                                       & $18\pm 3$                           & $25\pm 4$                              & 28            \\\hline
                            
\end{tabular}
\caption{Minimum number of data points needed to discover the equations for Eureqa and TuringBot tools vs Eureqa and TuringBot with \toolname Dataset. Each row shows the average min number of points required as well as the range over 5 separate trials. Results show a consistent reduction in points when using \toolname with some equations being discovered only when \toolname is used. {\bf Abbreviations Used}:- NoDisc: Does not Discover Equation. Disc: Enables Discovery. U: Unable to Run. \label{table:mainexp}}
\end{table*}

\begin{algorithm}[t]
\SetAlgoLined
\KwData{Input Dataset ${\bf X}, {\bf y}$}
\KwIn{Auxiliary Truth Set $T$, Alphabet $\Sigma$}
\KwResult{Predicted Equation $\widehat{f}$ and augmented dataset ${\bf X}, {\bf y}$}
 population P:= initial random set of equations\; 
 \While{generations $<$ numGens}{
    Calculate the total loss (MSE + Truth Error) for every equation in the population P over the dataset ${\bf X}, {\bf y}$
    
    Select the best equations from this list to survive and eliminate the rest 
    
    bestPerformer $\widehat{f}$ := best equation from this list 
    
    $\bf{X_{temp},y_{temp}}$ := data points for which B violates some truth in $T$ 
  
    Append $\bf{X_{temp}},\bf{y_{temp}}$ to dataset $\textbf{X}$,$\textbf{y}$
    
    Perform genetic mutation and crossover randomly on the remaining equations 
    
    P := new mutated and crossed over set of equations 
    
    Create a new randomly created set of equations and add it to the population
 }
 \Return augmented dataset \textbf{X},\textbf{y}, and predicted eqn $\widehat{f}$ 
 \caption{\toolname}
  \label{alg:LGGA}
\end{algorithm}

In this section, we describe our \toolname method described in Algorithm~\ref{alg:LGGA} (with its architecture described in Figure~\ref{fig:arch}). Broadly speaking, there are two primary modifications that we make to a classic GA. Please refer to the excellent work~\cite{koza1992genetic} for more details on classic GA.

\noindent{\textbf{Auxiliary Truth Enhanced Loss Function:}} Traditionally, the loss function of a GA is  MSE or some other metric to determine how well the current learnt function fits the provided dataset. By contrast, the AT Enhanced Loss Function is a weighted sum of MSE and {\it Truth Error} -- a metric which is higher for equations which have a higher degree of violation of the input ATs. To do this, we use a violation function $v_{t}(\widehat{f}, {\bf x})$ which is a measure of the violation of AT $t$ for a given data point ${\bf x}$ when using a candidate equation $\widehat{f}$. If $v_{t}(\widehat{f}, {\bf x}) > v_{t}(\overline{f}, {\bf x})$ implies that $\widehat{f}$ violates the AT more on the given datapoint than $\overline{f}$ does. We define Truth Error due to a single candidate function $\widehat{f}$ produced by an SR system as
\begin{equation}
    \mathrm{TruthError}(\widehat{f},T,X) := \frac{1}{|T|}\sum_{t\in T}\underset{{\bf x}\in \textbf{X}}{\mathrm{max}}(v_{t}(\widehat{f}, {\bf x}))
    \label{eqn:trutherror}
\end{equation}
\noindent where $T$ is the set of all ATs known a priori of $f$, ${\bf X}$ is dataset and $v_{t}$ is the violation function for truth $t\in T$.
\medskip 

Let us revisit the $R = \frac{r_1r_2}{r_1+r_2}$ example to show how this enhanced loss function works for the candidate equation: $\widehat{R} = r_1 + r_2$. Consider the dataset ${\bf X} = ((12, 0),(800, 0))$ and respective labels ${\bf y} = (0, 0)$. Using the AT $t: r_2=0\Rightarrow R=0$, we can define the violation function by $v_{t}(\widehat{R}, {\bf{x}}) = |\widehat{R}(r_1, 0)|$, for ${\bf{x_1}}$ this is $v_{t}(\widehat{R}, {\bf{x_1}}) = |\widehat{R}(12, 0)| = 12$. Notice how in the above example the violation $v_{t}(\widehat{R}, {\bf{x_2}}) = 800$ is a much larger degree of violation than $v_{t}(\widehat{R}, {\bf{x_1}})$, since the total Truth Error is the maximum of the individual violations the Truth Error for the above equation on that single auxiliary truth would be 800. The Truth Error for the equation over all ATs is just the sum of the Truth Error for every auxiliary truth we use for the equation, and since we have only one AT for this equation, its Truth Error is 800. This is captured in the final equation given the Auxiliary Truths and dataset ${\bf{X}}$ = $\{{\bf{x_1}}=(x_{1}^{1}, x_{2}^{1},..x_{n}^{1}), {\bf{x_2}}=(x_{1}^{2}, x_{2}^{2},.. x_{n}^{n}), ...\}$ given in equation~\ref{eqn:trutherror}.

\vspace{0.1cm} 
\noindent{\textbf{How \toolname Works:}} 
As described above in Algorithm~\ref{alg:LGGA}, LGGA takes in as input a dataset ${\bf X,y}$, a set of ATs $T$, and a set $\Sigma$ of allowed symbols. The output of \toolname is an augmented dataset ${\bf X_{\mathrm{aug}}}, {\bf y_{\mathrm{aug}}}$ and a proposed equation $\widehat{f}$. Like any classic GA, we begin with an unaugmented dataset and population of equations (line 1). However, unlike in classic GA, instead of having a fixed input dataset, the \toolname system progressively generates and adds {\it interesting} data points as the training goes on. Every time a new generation is created, we evaluate the AT Enhanced Loss described above in equation~\ref{eqn:trutherror}) of these candidate functions using the current training dataset and find the best performing function (lines 4, 5). We then use all the points in the current dataset to check whether any AT is violated for the equation of the best performer (line 6). All of these points are added back to our dataset (line 7). This continues until the specified generation limit is met, or the overall error reaches a threshold. 

\noindent{\textbf{Value Addition of Data Augmentation:}} The use of ATs is very powerful because it often allows us to generate completely new data points without having to query an oracle to generate useful new data, where an oracle is a system that takes in a data point ${\bf x \in X}$ and returns a label $f({\bf x}$. Access to an oracle implies that we can continuously query the target equation throughout.   

To illustrate this, take an example where we try to learn an unknown function $f$ where the given AT is that $f$ is symmetric in its arguments. We will illustrate the data augmentation process for candidate equation $\widehat{f}(x_1, x_2)$,  If we were provided the data point $f(2, 3) = 4$ then to evaluate whether $\widehat{f}$ is compliant with the AT we would check if $|\widehat{f}(2, 3) - \widehat{f}(3, 2)| > 0$. However, in doing this we realize that we can now create a \textbf{new data point for free} which is $f(3, 2) = 4$. We are guaranteed that this is an \textbf{sound} datapoint by the auxiliary truth, and it is possibly a \textbf{non-trivial} data point as a record of $f(3, 2)$ may not exist in the dataset we currently possess. In doing so, we not only augment the dataset by adding new and distinct data points but also data points that we know the best equation currently classifies incorrectly and hence would likely have a guiding effect on the model. 

\section{Experimental Results}
\label{sec:exp}
In this section, we describe the experiments we ran to test the efficacy of our \toolname tool. All experiments below were run on a system with an Intel(R) Core(TM) i7-8550U CPU @ 1.80GHz, 16GB RAM.  
\vspace{0.1cm}

\noindent{\textbf{Experiment 1: Classic GA vs LGGA.}} In this experiment, we compared a classic GA algorithm (DEAP symbolic regressor~\cite{DEAP_JMLR2012}) against \toolname (an LGGA enhanced version of the DEAP symbolic regression) and assessed them based on how many equations from the Feynman book \cite{feynman2011feynman} they were able to learn within a time limit. We used the equations of Table~\ref{table:equation_list} as our benchmark and their associated ATs. For each of these equations, we followed the process detailed below:

\begin{enumerate}
    \item Run the \toolname on a random dataset of initial size $m = 100$, terminate if the MSE of the best equation is below a specified $\epsilon = 10^{-4}$ threshold (or if the algorithm exceeds a certain generation time out). Upon termination the tool returns an augmented dataset of size $m^{*} = 100 + k$ (where k is the number of new data points LGGA has created) and an equation prediction

    \item Run the classic GA with a random dataset of fixed size $m^{*}$ and terminate under the same conditions. The reason we use the same amount of data for each tool $m^{*}$ is we wish to show that the value addition from \toolname comes not just from the fact that we have more data, but also that the AT enhanced feedback loop and data augmentation have a significant guiding impact

    \item Compare them on the metric of how many equations they were able to solve, where we define solve as terminating with an equation that is semantically equivalent to the target equation
\end{enumerate}

\noindent We conducted the above experiment 15 times with different random seeds. If a method discovers the equation at least once out of these 15 runs, we declare the equation solved. 

\vspace{0.1cm}
\noindent{\textbf{Classic GA vs. \toolname Result:}}
The \toolname dominates the classic approach. Importantly, given exactly the same amount of data, \toolname discovers 5 out of 16 equations (Snell, Coulomb, Resistance, Reflection, Gas), none of which the classic GA can solve. For the equations that neither could solve, the MSE of the \toolname prediction can be up to 70\% less than that of the classic GA. This strengthens our claim that the paper's two novel additions - the AT Enhanced Loss Function and Data Augmentation, do have a significant guiding impact.

\noindent{\textbf{Experiment 2: Industrial SR tools vs. the corresponding LGGA-enhanced Versions.}}
In this experiment, we test \toolname's ability to generate richer and more informative datasets to augment industrial-strength SR tools and quantify how effective this data augmentation procedure is in improving the highly-engineered SR tools Eureqa and TuringBot. To do this, We used all the equations and truths in Table 1. For each of the 16 equations, with both tools - Eureqa and TuringBot, we follow the process below:

\begin{enumerate}
    \item {\bf Execution of Data Augmented Eureqa and TuringBot:} Supply \toolname with the a random initial dataset ${\bf X,y}$ of size $m$, alphabet of allowed symbols $\Sigma$, and ATs $T$ to produce an augmented dataset ${\bf X_{\mathrm{aug}, m'}}$ of size $m'$. Run Eureqa (resp. TuringBot) with a timeout of 15 minutes; if within these 15 minutes, it proposes an equation that is semantically equivalent to the target expression, we say that Eureqa has solved the equation with $m'$ data points. We repeat this process and generate a completely new augmented dataset ${\bf X_{\mathrm{aug}, m'-1}}$ of size $m'-1, m'-2 .. etc$, steadily lowering the number of data points until we reach a number $m'_{\mathrm{LGGA}}$ such that Eureqa (resp. TuringBot) can solve the equation with $m'_{\mathrm{LGGA}}$ augmented data points but cannot solve the equation with $m'_{\mathrm{LGGA}}-1$ data points. We call this the minimum data points from the augmented dataset that Eureqa (resp. TuringBot) needs to solve. Notice that we do not consider the initial dataset size $m$ to be the minimum points needed ($m < m'_{\mathrm{LGGA}}$ always) even though the rest of the points were computed by \toolname without an oracle labeller. This is because we wish to show that the value addition of \toolname is not only that it can create more data, but also more informative data. 
    
    \item {\bf Execution of Eureqa (resp. TuringBot) without Data Augmentation:} Produce a random dataset ${\bf X_{\mathrm{rand}}}$ of size $m''$. We follow the same procedure above to reach an $m''_{\mathrm{RAND}}$ such that Eureqa (resp. TuringBot) can solve the equation with $m''_{\mathrm{RAND}}$ random data points but cannot solve the equation with $m''_{\mathrm{RAND}}-1$ data points. We call this the minimum data points from the random dataset that Eureqa (resp. TuringBot) needs to solve the equation.
    
    \item {\bf Evaluation: }We then use as a metric the improvement in data efficiency: $\frac{m''_{\mathrm{RAND}}-m'_{\mathrm{LGGA}}}{m''_{\mathrm{RAND}}}$\% as a way to measure the reduction \% of data needed, in an apple-to-apple comparison of Eureqa vs. Data Augmented Eureqa (similarly, for TuringBot)
\end{enumerate}

\noindent The above experiments were repeated five times for each equation, and we noted the mean and deviation of the minimum number of data points needed to learn each equation.

\noindent{\textbf{Results 2:}}
The Eureqa tool shows significant improvements when used with an \toolname augmented dataset (shown in Table~\ref{table:mainexp}). Every equation shows a reduction in the minimum data points needed for discovery, and three equations (namely, reflection, distance, and normal) are discovered only when \toolname augmented data is used. We were not given access to the full version of Eureqa and could not test our tool on equations with six or more input variables (Marked as U for Unable to Run).

TuringBot with \toolname is the most powerful tool out of all of those we tested (except for cases where there is a square root in the equation as TuringBot does not have a square root operator). Even with such a highly engineered tool as TuringBot, we see a consistent reduction in the number of data points needed as seen in Table~\ref{table:mainexp}.

This experiment shows that \toolname has a significant positive influence on the data efficiency and convergence rates of industrial SR tools. The average time taken for \toolname to generate the dataset was only $4.34 \pm 1.5s$. 

\noindent{\textbf{Experiment 3: AI Feynman}} In this experiment, we both attempted to test our \toolname tool in conjunction with AI Feynman and also attempted to test AI Feynman against our Eureqa + \toolname results from table~\ref{table:mainexp}. The authors of AI Feynman 1.0 report that their tool discovers 30\% more equations from the Feynman book compared to Eureqa~\cite{udrescu2020ai}. However, AI Feynman 1.0 is very sensitive to noise, so they developed the AI Feynman 2.0 \cite{udrescu2020ai2}, which is far more robust at the cost of significantly longer runtimes. Unfortunately, version 1.0 is not available, and we had to conduct our experiments with AI Feynman 2.0.  We spent dozens of hours on AI Feynman 2.0 and communicated with the author as well. Unfortunately, we were not able to get AI Feynman 2.0 to terminate on any of the equations in the table~\ref{table:equation_list} within a timeout of 3 hours (the timeout we used for the other tools was 15 minutes).

\section{Analysis and Limitations}
\label{sec:analysis}

\noindent\textbf{Encoding Boundary Points}: Experiments \#1, \#2, demonstrate that SR tools using \toolname's augmented datasets outperform random datasets. This result empirically demonstrates that new data points produced from ATs can be significantly more informative than just having additional random data. The way a function behaves around key points of interest, including domain boundaries, strongly defines the function's behavior in general. Thus, augmenting a dataset with these points significantly reduces the set of viable choices of expressions that an SR system has to make. Since we only do formula evaluation during the checking of ATs this process ends up being very efficient as well, as seen in experiment \#2.

\noindent\textbf{Deterring Overfitting:} Recall Eureqa's incorrect prediction of the gas equation in equation~\ref{eq:eureqa_wrong} when a very small input dataset is given. This equation was Eureqa's final prediction, generated while running experiment \#2, which provides strong evidence that the augmented dataset discourages models from producing complicated overfitted equations. Most modern SR techniques consider both expression complexity (i.e. number of operations used) and fitting error (e.g. MSE) when evaluating expressions. These models usually avoid overfitting by penalizing high complexity expressions, allowing them only when they admit a low MSE. Regularization is less effective with small datasets since overfitted equations can achieve a near-zero MSE when data is scarce. These overfitting expressions will be consistently favored for future generations, even if they are very complex. We correct this issue in \toolname by replacing the traditional loss function with a weighted combination AT-enhanced and MSE loss function. An additional corrective measure we take in \toolname is the data augmentation method. 

\noindent\textbf{Dataset Curation}: Since \toolname only produces new data points for ATs that are violated, it naturally focuses on the ATs that are harder to satisfy. This means scientists can provide multiple ATs of varying strengths and rely on the \toolname feedback loop to prioritize the stronger, more informative truths. This provides a curated dataset that is more efficient than treating all ATs equally and generating counterexamples for each of them. To verify this, we started with a random dataset and manually produced new data points for every AT. This method not only scaled poorly when there were many ATs but also showed far inferior results when compared to an \toolname dataset of the same size. This shows that \toolname's online feedback loop is effective at picking the most useful ATs to generate counterexamples for and curating the dataset in a more optimal way than simply producing new data from ATs in a single pre-processing step. 

\noindent\textbf{Dependence on ATs:} The most explicit limitation of \toolname is its dependence on informative ATs. This requirement means that before predicting any equation, the SR user must put in some effort into a robust preliminary data analysis to come up with ATs.

\noindent{\textbf{Lack of Proof:}}The \toolname does not use a rigorous proof to see whether a candidate equation violates a constraint, but checks it on the data points in the current dataset. Thus, an equation may violate a constraint, but \toolname is not able to detect it. One solution is to use an SMT solver which takes in an equation and, through a formal proof, determines whether an equation over its \textbf{entire domain} adheres to a constraint or not. 

\noindent{\textbf{Limited Diversity of new points:}} Every new data point generated by \toolname must be some transformation of an existing data point using an AT that the candidate equation violates. Restricting the new data points in this way has the benefit that \toolname does not require access to an oracle, which can be expensive, but comes at the cost of diversity of points generated. 

\noindent{\textbf{Simple GA:}} Lastly, the current version of the \toolname likely produces a less than optimal dataset, since we are using a simple GA. We expect \toolname to perform even better if we replace the classic GA in the \toolname loop (See Figure~\ref{fig:arch}) with highly-engineered systems such as TuringBot or Eureqa. We are currently constrained by the fact these systems are closed source and were denied access when we reached out to them.


\section{Related Work}
\label{sec:related}
\textbf{Symbolic Regression:} We refer the reader to the following papers~\cite{keijzer2003improving,karaboga2012artificial,kim2020integration} on recent advances in SR. The closest of these works to \toolname is Kubalik et al. where they discuss the applicability of ATs as part of the loss function in GAs~\cite{kubalik2020symbolic}. However, our approach, by contrast, not only uses ATs as part of the loss function, but also as part of data augmentation.  

\noindent
\textbf{Combinations of Logic and Machine Learning:} There is a vast literature on combinations of logic and machine learning, and we refer the reader to the following excellent survey~\cite{amel2019shallow}. Of these, the work most relevant to \toolname is logic guided machine learning (LGML)~\cite{scott2020lgml}. To the best of our knowledge, it is one of the early attempts at using ATs to produce richer datasets in an online fashion, via the use of an SMT solver, with the aim of making SR tools more efficient. The LGML tool is different from \toolname in two ways. First, LGML uses an SMT solver to find counterexample between candidate learnt functions and ATs, at the cost of solving hard satisfiability problems. By contrast, \toolname performs consistency checks on boundary values and existing dataset via formula evaluation, a considerably cheaper option. Of course, the method used by \toolname is also weaker as a consequence, but good enough for the settings we considered. Second, LGML necessarily requires access to an oracle which can produce labels for counterexamples that the underlying SMT solver finds. By contrast, \toolname does not require an oracle.

\noindent{\textbf{Active Learning:}} The idea of online data augmentation itself is well known, particularly in image-related tasks, for improving the utility of training data based on transformations on existing data points. Some developments in this area include performing transformations on a learned feature space~\cite{devries2017dataset}, and directly learning efficient augmentation strategies~\cite{Cubuk_2019_CVPR}. Unlike these approaches, our method performs augmentation by making use of not only the original data but also ATs.

\section{Conclusions and Future Work }
\label{sec:conclusion}
In this paper, we presented \toolname, a constraint-enhanced GA that uses ATs to drive data augmentation and compute loss. Combining \toolname with SR tools results in solving up to \uniquesolved more equations and up to a \dataefficency improvement in data efficiency. In the future, we plan to combine \toolname with an SMT Solver in a way similar to LGML and using it occasionally to produce more useful data. We also could look to use the fact that AI Feynman is open source and integrate the \toolname loop into its mechanism. Finally, we could try to use the AT Enhanced Loss Function and data generation techniques with industrial grade SR solutions, to see if we could make an end to end predictor that on its own can beat the state-of-the-art.

\bibliography{main.bib}

\begin{thebibliography}{22}
\providecommand{\natexlab}[1]{#1}
\providecommand{\url}[1]{\texttt{#1}}
\providecommand{\urlprefix}{URL }
\expandafter\ifx\csname urlstyle\endcsname\relax
  \providecommand{\doi}[1]{doi:\discretionary{}{}{}#1}\else
  \providecommand{\doi}{doi:\discretionary{}{}{}\begingroup
  \urlstyle{rm}\Url}\fi

\bibitem[{Amel(2019)}]{amel2019shallow}
Amel, K.~R. 2019.
\newblock From shallow to deep interactions between knowledge representation,
  reasoning and machine learning.
\newblock In \emph{Proceedings 13th International Conference Scala Uncertainity
  Mgmt (SUM 2019), Compi{\`e}gne, LNCS}, 16--18.

\bibitem[{Belle(2020)}]{belle2020symbolic}
Belle, V. 2020.
\newblock Symbolic Logic meets Machine Learning: A Brief Survey in Infinite
  Domains.
\newblock \emph{arXiv preprint arXiv:2006.08480} .

\bibitem[{Billard and Diday(2002)}]{billard2002symbolic}
Billard, L.; and Diday, E. 2002.
\newblock Symbolic regression analysis.
\newblock In \emph{Classification, Clustering, and Data Analysis}, 281--288.
  Springer.

\bibitem[{Cubuk et~al.(2019)Cubuk, Zoph, Mane, Vasudevan, and
  Le}]{Cubuk_2019_CVPR}
Cubuk, E.~D.; Zoph, B.; Mane, D.; Vasudevan, V.; and Le, Q.~V. 2019.
\newblock AutoAugment: Learning Augmentation Strategies From Data.
\newblock In \emph{Proceedings of the IEEE/CVF Conference on Computer Vision
  and Pattern Recognition (CVPR)}.

\bibitem[{DeVries and Taylor(2017)}]{devries2017dataset}
DeVries, T.; and Taylor, G.~W. 2017.
\newblock Dataset augmentation in feature space.
\newblock \emph{arXiv preprint arXiv:1702.05538} .

\bibitem[{Dominic, Leahy, and Willis(2010)}]{gptips}
Dominic, P.; Leahy, D.; and Willis, M. 2010.
\newblock \emph{Predicting the toxicity of chemical compounds using GPTIPS: A
  free genetic programming toolbox for MATLAB}, volume~70, 83--93.
\newblock Springer.
\newblock \doi{10.1007/978-94-007-0286-8_8}.

\bibitem[{Feynman, Leighton, and Sands(2011)}]{feynman2011feynman}
Feynman, R.~P.; Leighton, R.~B.; and Sands, M. 2011.
\newblock \emph{The Feynman lectures on physics, Vol. I: The new millennium
  edition: mainly mechanics, radiation, and heat}, volume~1.
\newblock Basic books.

\bibitem[{Fortin et~al.(2012)Fortin, {De Rainville}, Gardner, Parizeau, and
  Gagn\'e}]{DEAP_JMLR2012}
Fortin, F.-A.; {De Rainville}, F.-M.; Gardner, M.-A.; Parizeau, M.; and
  Gagn\'e, C. 2012.
\newblock {DEAP}: Evolutionary Algorithms Made Easy.
\newblock \emph{Journal of Machine Learning Research} 13: 2171--2175.

\bibitem[{{Horn}, {Nafpliotis}, and {Goldberg}(1994)}]{nicheparetogp}
{Horn}, J.; {Nafpliotis}, N.; and {Goldberg}, D.~E. 1994.
\newblock A Niched Pareto Genetic Algorithm for Multiobjective Optimization.
\newblock In \emph{Proceedings of the First IEEE Conference on Evolutionary
  Computation. IEEE World Congress on Computational Intelligence}, 82--87
  vol.1.

\bibitem[{Karaboga et~al.(2012)Karaboga, Ozturk, Karaboga, and
  Gorkemli}]{karaboga2012artificial}
Karaboga, D.; Ozturk, C.; Karaboga, N.; and Gorkemli, B. 2012.
\newblock Artificial bee colony programming for symbolic regression.
\newblock \emph{Information Sciences} 209: 1--15.

\bibitem[{Keijzer(2003)}]{keijzer2003improving}
Keijzer, M. 2003.
\newblock Improving symbolic regression with interval arithmetic and linear
  scaling.
\newblock In \emph{European Conference on Genetic Programming}, 70--82.
  Springer.

\bibitem[{Kim et~al.(2020)Kim, Lu, Mukherjee, Gilbert, Jing, {\v{C}}eperi{\'c},
  and Solja{\v{c}}i{\'c}}]{kim2020integration}
Kim, S.; Lu, P.~Y.; Mukherjee, S.; Gilbert, M.; Jing, L.; {\v{C}}eperi{\'c},
  V.; and Solja{\v{c}}i{\'c}, M. 2020.
\newblock Integration of Neural Network-Based Symbolic Regression in Deep
  Learning for Scientific Discovery.
\newblock \emph{IEEE Transactions on Neural Networks and Learning Systems} .

\bibitem[{Koza and Koza(1992)}]{koza1992genetic}
Koza, J.~R.; and Koza, J.~R. 1992.
\newblock \emph{Genetic programming: on the programming of computers by means
  of natural selection}, volume~1.
\newblock MIT press.

\bibitem[{Kubal{\'\i}k, Derner, and Babu{\v{s}}ka(2020)}]{kubalik2020symbolic}
Kubal{\'\i}k, J.; Derner, E.; and Babu{\v{s}}ka, R. 2020.
\newblock Symbolic Regression Driven by Training Data and Prior Knowledge.
\newblock \emph{arXiv preprint arXiv:2004.11947} .

\bibitem[{Lu, Ren, and Wang(2016)}]{lu2016using}
Lu, Q.; Ren, J.; and Wang, Z. 2016.
\newblock Using genetic programming with prior formula knowledge to solve
  symbolic regression problem.
\newblock \emph{Computational intelligence and neuroscience} 2016.

\bibitem[{Pal and Wang(1996)}]{pal1996genetic}
Pal, S.~K.; and Wang, P.~P. 1996.
\newblock \emph{Genetic algorithms for pattern recognition}.
\newblock CRC press.

\bibitem[{Schmidt and Lipson(2009)}]{schmidt2009distilling}
Schmidt, M.; and Lipson, H. 2009.
\newblock Distilling free-form natural laws from experimental data.
\newblock \emph{science} 324(5923): 81--85.

\bibitem[{Scott, Panju, and Ganesh(2020)}]{scott2020lgml}
Scott, J.; Panju, M.; and Ganesh, V. 2020.
\newblock LGML: Logic Guided Machine Learning (Student Abstract).
\newblock In \emph{Proceedings of the AAAI Conference on Artificial
  Intelligence}, 10, 13909--13910.

\bibitem[{Towfighi(2020)}]{towfighi2020symbolic}
Towfighi, S. 2020.
\newblock Symbolic regression by uniform random global search.
\newblock \emph{SN Applied Sciences} 2(1): 34.

\bibitem[{TuringBot(2020)}]{turingbot}
TuringBot, S. 2020.
\newblock Symbolic Regression Software.
\newblock \urlprefix\url{https://turingbotsoftware.com/}.

\bibitem[{Udrescu et~al.(2020)Udrescu, Tan, Feng, Neto, Wu, and
  Tegmark}]{udrescu2020ai2}
Udrescu, S.-M.; Tan, A.; Feng, J.; Neto, O.; Wu, T.; and Tegmark, M. 2020.
\newblock AI Feynman 2.0: Pareto-optimal symbolic regression exploiting graph
  modularity.
\newblock \emph{arXiv preprint arXiv:2006.10782} .

\bibitem[{Udrescu and Tegmark(2020)}]{udrescu2020ai}
Udrescu, S.-M.; and Tegmark, M. 2020.
\newblock AI Feynman: A physics-inspired method for symbolic regression.
\newblock \emph{Science Advances} 6(16): eaay2631.

\end{thebibliography}

\end{document}